\documentclass[11pt]{article}
\usepackage{color}
\usepackage{algorithm, algpseudocode}
\usepackage{mathrsfs}
\usepackage{dsfont}
\usepackage{lmodern}
\usepackage{array}
\usepackage{bbm}
\usepackage{amsmath}
\usepackage{amssymb}
\usepackage[mathscr]{eucal}
\usepackage{graphicx}
\usepackage{mathrsfs}
\usepackage{psfrag}
\usepackage{color}
\usepackage{here}
\usepackage[margin=1in]{geometry}
\usepackage{wasysym}
\usepackage{enumitem}
\usepackage{xcolor}
\usepackage{caption}
\usepackage{subcaption}
\usepackage{amsthm}
\usepackage{lipsum}
\usepackage[round]{natbib}

\definecolor{darkblue}{RGB}{0,0,140}
\usepackage[colorlinks=true,
            allcolors=darkblue]{hyperref}

\newtheorem{theorem} {Theorem}
\newtheorem{conjecture} {Conjecture}
\newtheorem{proposition} {Proposition}
\newtheorem{lemma} {Lemma}

\newcommand{\Exp}{\mathds{E}}

\newcommand{\Prob}{\mathds{P}}

\newcommand{\Nat}{\mathbb{N}}

\DeclareMathOperator*{\argmax}{arg\,max}

\newcommand{\Fc}{\mathcal{F}}

\newcommand{\Sc}{\mathcal{S}}
\newcommand{\Ac}{\mathcal{A}}

\newcommand{\Hc}{\mathcal{H}}

\title{On Lower Bounds for Regret in Reinforcement Learning}

\author{
Ian Osband \\
Stanford University, Google DeepMind\\
\texttt{iosband@stanford.edu}
\and
Benjamin Van Roy \\
Stanford University\\
\texttt{bvr@stanford.edu}
}
\begin{document}
\maketitle

\section{Introduction}

This is a brief technical note to clarify the state of lower bounds on regret for reinforcement learning.
In particular, this paper:
\begin{itemize}
     \item Reproduces a lower bound on regret for reinforcement learning, similar to the result of Theorem 5 in the journal UCRL2 paper \cite{Jaksch2010}.

     \item Clarifies that the proposed proof of Theorem 6 in the REGAL paper \cite{Bartlett2009} does not hold using the standard techniques without further work. We suggest that this result should instead be considered a conjecture as it has no rigorous proof.

     \item Suggests that the conjectured lower bound given by \cite{Bartlett2009} is incorrect and, in fact, it is possible to improve the scaling of the upper bound to match the weaker lower bounds presented in this paper.

\end{itemize}

\section{Problem formulation}
\label{sec: prob form}

We consider the problem of learning to optimize an unknown MDP $M^* = (\Sc, \Ac, R^*, P^*)$.
$\Sc = \{1,..,S\}$ is the state space, $\Ac=\{1,..,A\}$ is the action space.
In each timestep $t=1,2,..$ the agent observes a state $s_t \in \Sc$, selects an action $a_t \in \Ac$, receives a reward $r_t \sim R^*(s_t,a_t) \in [0,1]$ and transitions to a new state $s_{t+1} \sim P^*(s_t, a_t)$.
We define all random variables with respect to a probability space $(\Omega, \Fc, \Prob)$.

A policy $\mu$ is a mapping from state $s \in \Sc$ to action $a \in \Ac$.
For MDP $M$ and any policy $\mu$ we define the long run average reward starting from state $s$:
\begin{equation}
    \lambda^M_\mu(s) := \lim_{T \rightarrow \infty}
    \Exp_{M, \mu} \left[ \frac{1}{T} \sum_{t=1}^T \overline{r}(s_t,a_t) \ \mid \ s_1 = s \right],
\end{equation}
where $\overline{r}^*(s,a) := \Exp[r | r \sim R^*(s,a)]$.
The subscripts $M, \mu$ indicate the MDP evolves under $M$ with policy $\mu$.
A policy $\mu^M$ is optimal for the MDP $M$ if $\mu^M \in \arg\max_\mu \lambda^M_\mu(s)$ for all $s \in \Sc$.
For the unknown MDP $M^*$ we will often abbreviate sub/superscripts to simply $*$, for example $\lambda^*_*$ for $\lambda^{M^*}_{\mu^{M^*}}$.

Let $\Hc_t=(s_1, a_1, r_1,..,s_{t-1}, a_{t-1}, r_{t-1})$ denote the history of observations made \textit{prior} to time $t$.
A reinforcement learning algorithm is a deterministic sequence $\{ \pi_t | t=1,2,.. \}$ of functions each mapping $\Hc_t$ to a probability distribution $\pi_t(\Hc_t)$ over policies, from which the agent sample policy $\mu_t$ at timestep $t$.
We define the regret of a reinforcement learning algorithm $\pi$ up to time $T$
\begin{equation}
    {\rm Regret}(T, \pi, M^*)(s) := \sum_{t=1}^T \left\{ \lambda^*_*(s) - r_t \right\} \bigg| s_1 = s.
\end{equation}

The regret of a learning algorithm shows how worse the policy performs that optimal in terms of cumulative rewards.
Any algorithm with $o(T)$ regret will \textit{eventually} learn the optimal policy.
Note that the regret is random since it depends on the unknown MDP $M^*$, the random sampling of policies and, through the history $\Hc_t$ on the previous transitions and rewards.
We will assess and compare algorithm performance in terms of the regret.

\subsection{Finite horizon MDPs}

We now spend a little time to relate the formulation above to so-called finite horizon MDPs \cite{Osband2013,dann2015sample}.
In this setting, an agent will interact repeatedly with a environment over $H \in \Nat$ timesteps which we call an \textit{episode}.
A finite horizon MDP $M^* = (\Sc, \Ac, R^*, P^*, H, \rho)$ is defined as above, but every $H \in \Nat$ timesteps the state will reset according to some initial distribution $\rho$.
We call $H \in \Nat$ the horizon of the MDP.

In a finite horizon MDP a typical policy may depend on both the state $s \in \Sc$ and the timestep $h$ within the episode.
To be explicit, we define a policy $\mu$ is a mapping from state $s \in \Sc$ and period $h=1,..,H$ to action $a \in \Ac$.
For each MDP $M = (\Sc, \Ac, R^M \hspace{-1mm}, P^M\hspace{-1mm}, H, \rho)$ and policy $\mu$ we define the state-action value function for each period $h$:
\vspace{-1mm}
\begin{equation}
\label{eq: q value tabular}
  Q^{M}_{\mu, h}(s, a) := \Exp_{M,\mu}\left[ \sum_{j=h}^{H} \overline{r}^M(s_j,a_j) \Big| s_h = s, a_h=a \right],
\end{equation}
and $V^{M}_{\mu, h}(s) := Q^M_{\mu, h}(s, \mu(s,h))$.
Once again, we say a policy $\mu^M$ is optimal for the MDP $M$ if $\mu^M \in \argmax_{\mu} V^{M}_{\mu, h}(s)$ for all $s \in \Sc$ and $h=1,\ldots,H$.

At first glance this might seem at odds with the formulation in Section \ref{sec: prob form}.
However, finite horizon MDPs can be thought of as a special case of Section \ref{sec: prob form} in the expanded state space $\tilde{\Sc} := \Sc \times \{1,..,H\}$.
In this case it is typical to assume that the agent \textit{knows} about the evolution of time $h$ deterministically a priori.
To highlight this time evolution within episodes, with some abuse of notation, we let $s_{kh} = s_t$ for $t=(k-1)H+h$, so that
$s_{kh}$ is the state in period $h$ of episode $k$.  We define $\Hc_{kh}$ analogously.

\section{Multi-armed bandit}
\label{sec: mab}

We call the degenerate MDP with only one state $S=1$ a multi-armed bandit with independent arms \cite{lai1985asymptotically}.
In this setting the actions $a_t \in \Ac$ are often called ``arms'' and the optimal average reward is simply the average reward of the highest reward,
$$ \lambda^*_* = \max_{a} \overline{r}^*(a).$$
We now reproduce a lower bound on regret for any learning algorithm in a multi-armed bandit \cite{Bubeck2012regretBandit}.

\begin{theorem}[Lower bound on regret in bandits]
\label{thm: bandit_lower}
\hspace{0.000001mm} \newline
Let $\sup$ be the supremum over all distributions of rewards such that for each $a=1,..,A$ the rewards $r(1)_t, .., r(A)_t \in \{0,1\}$ are i.i.d. and let $\inf$ be the infimum over all reinforcement learning algorithms.
Then
\begin{equation}
    \inf \sup \left( \max_a \overline{r}^*(a) T - \Exp \left[ \sum_{t=1}^T  \overline{r}^*(a_t) \right]\right) \ge \frac{1}{24} \sqrt{AT}.
\end{equation}

\end{theorem}

At a high level Theorem \ref{thm: bandit_lower} says that no matter what learning algorithm you choose, there will always be some environment which gives your algorithm $\Omega(\sqrt{AT})$ regret.
This is a pretty powerful result, since it means that if we can design an algorithm with upper bounds on regret $O(\sqrt{AT})$ then this algorithm is in some sense near-optimal \cite{Bubeck2012regretBandit}.

The intuition for the proof is relatively simple and presented in \cite{Bubeck2012regretBandit}.
After any $T$ timesteps there must be some arm which is pulled less than $T / A$ times.
Standard concentration results state that the estimates of a random variable can only be accurate up to $O\left( \frac{1}{\sqrt{n}}\right)$ where $n$ is the number of observations.
Therefore, for the arm with $n \le T / A$ it is difficult to distinguish between a ${\rm Ber}(1/2)$ and ${\rm Ber}(1/2 + \sqrt{A/ T})$.
This means that, if every arm is ${\rm Ber}(1/2)$ but one ${\rm Ber}(1/2 + \sqrt{A/T})$, any algorithm would incur $T \sqrt{A / T} = \sqrt{AT}$ regret.
In the next section we will see how to make this argument more rigorous.

\subsection{Proof of Theorem \ref{thm: bandit_lower}}

We consider the problem where all arms are i.i.d. Bernoulli with parameter $\delta$, but one arm $a^*$ has parameter $\delta + \epsilon$ for some $\delta, \epsilon > 0$.
We define an auxilliary $\tilde{r}_t(a) = r_t(a)$ for all $a \neq a^*$, but with the rewards of the action $a=a^*$ replaced by the draw $\tilde{r}_t \sim {\rm Ber}(\delta)$.
We consider an auxilliary sequence of actions $\tilde{a}_t \sim \pi_t(\tilde{H}_t)$ for $\tilde{H}_t = (\tilde{a}_1, \tilde{r}_1,.., \tilde{a}_{t-1}, \tilde{r}_{t-1})$ as the history generated by an agent with no feedback informing them about $a^*$.

We introduce the notation $n_T(a) := | \{a_t = a | t=1,..,T\} |$ and $\tilde{n}_T(a) := | \{\tilde{a}_t = a | t=1,..,T\} |$ to denote the number of times arm $a$ have been selected by time $T$ under $a_t$ and $\tilde{a}_t$ respectively.
The following lemma establishes a lower bound on the regret realized by action $\tilde{a}_t$.

\begin{lemma}[Regret of an uninformed agent]
\label{lem: regret_uninformed}
\hspace{0.000001mm} \newline
For all $\delta, \epsilon > 0$ and all learning algorithms $\pi$,
$$ \max_a \overline{r}^*(a) T - \Exp \left[ \sum_{t=1}^T  \overline{r}^*(\tilde{a}_t) \right] \ge \frac{A-1}{A} T \epsilon.$$

\begin{proof}
We have,
\begin{eqnarray}
    \max_a \overline{r}^*(a) T - \Exp \left[ \sum_{t=1}^T  \overline{r}^*(\tilde{a}_t) \right]
    &=& \Exp \left[ \sum_{a \neq a^*} \tilde{n}_T(a) \epsilon \right] \nonumber \\
    &=& \epsilon(T - \tilde{n}_T(a^*)) \nonumber \\
    &=& \epsilon T (1 - \frac{1}{A} ),
\end{eqnarray}
where the last step follows from a symmetry argument, since $a^*$ is independent of $\tilde{n}_t(a)$ for all actions $a$.
\end{proof}
\end{lemma}

We now establish that, if $\epsilon$ is sufficiently small, then over a limited time horizon the distributions of $\tilde{r}_t(a_t)$ cannot be significantly different from the outcomes $r_t(a_t)$.
We compare the conditional distributions over the choice of action $P$ with the choice of actions $\tilde{P}$ which would have arisen under the uninformative data $\Hc_t$.
To be more precise we define $P(z_t^T | \Hc_t) := \Prob(r^T_t = z^T_t | \Hc_t) $ with $\tilde{P}(z_t^T | \Hc_t) := \Prob(\tilde{r}_t^T = z_t^T | \tilde{\Hc_t})$.
We write $r_t^T := (r_t(a_t),..,r_T(a_t))$ for the sequence of rewards from time $t$ to $T$ and similarly for $\tilde{r}_t^T$.
To quantify the difference between two distributions we will employ the following notion of KL divergence:
\begin{equation}
    d_{KL}\left( \tilde{P}(z_t^T| \tilde{H}_t) , P(z_t^T| H_t) ) \right)
    = \Exp \left[ \sum_{z_t^T}  \tilde{P}(z_t^T| \tilde{H}_t)
    \log \left( \frac{\tilde{P}(z_t^T| \tilde{H}_t) }{P(z_t^T| H_t)}\right) \right].
\end{equation}

\begin{lemma}[KL divergence of uninformed distribution]
\label{lem: kl_uninformed}
\hspace{0.000001mm} \newline
For all $\delta, \epsilon > 0$ and all learning algorithms $\pi$,
$$ d_{KL}\left( \tilde{P}(z_1^T| \tilde{H}_t) , P(z_1^T| H_t) ) \right) \le \frac{T}{A} \left( \delta \log \frac{\delta}{\delta + \epsilon} + (1 - \delta) \log \frac{1- \delta}{1 - \delta - \epsilon}\right).$$

\begin{proof}
We can apply the chain rule of KL divergence \cite{Bubeck2012regretBandit} to obtain
$$ d_{KL}\left( \tilde{P}(z_t^T| \tilde{H}_t) , P(z_t^T| H_t) ) \right)
 = \sum_{t=1}^T d_{KL}\left( \tilde{P}(z_t^t| \tilde{H}_t) , P(z_t^t| H_t) ) \right).$$
It follows that
\begin{eqnarray*}
    d_{KL}\left( \tilde{P}(z_t^T| \tilde{H}_t) , P(z_t^T| H_t) ) \right)
    &=& \sum_{t=1}^T \Prob(\tilde{a}_t \neq a^*) \left( \delta \log \frac{\delta}{\delta + \epsilon} + (1 - \delta) \log \frac{1- \delta}{1 - \delta - \epsilon} \right).
\end{eqnarray*}
We conclude the proof by noting that the actions $\tilde{a}_t$ are selected indepedently of of $a^*$ together with a symmetry argument.
\end{proof}
\end{lemma}

We now use Pinsker's inequality to show that, if the distribution of actions $P$ is close to the choice of actions under uninformative data $\tilde{P}$ then the resulting regret is close to the regret of the uninformative policy.

\begin{lemma}[Regret bound in terms of KL divergence]
\label{lem: regret_kl}
\hspace{0.000001mm} \newline
For all $\delta, \epsilon > 0$ and all learning algorithms $\pi$,
$$ \max_a \overline{r}^*(a) T - \Exp \left[ \sum_{t=1}^T  \overline{r}^*(a_t) \right] \ge  \epsilon T \left( 1 - \frac{1}{A} - \sqrt{\frac{1}{2} d_{KL}(\tilde{P}(z_1^T) , P(z_1^T))} \right). $$

\begin{proof}
Pinsker's inequality gives us
$$ \Exp \left[ \frac{n_T(a^*)}{T} - \frac{\tilde{n}_T(a^*)}{T} \right]
    \le \sqrt{\frac{1}{2} d_{KL}(\tilde{P}(z_1^T) , P(z_1^T ))}. $$
Since $\Exp[\tilde{n}_T(a^*) ] = T / A$, it follows that
$ \Exp \left[ \frac{n_T(a^*)}{T} \right]
    \le \sqrt{\frac{1}{2} d_{KL}(\tilde{P}(z_1^T) , P(z_1^T ))} + \frac{1}{A}. $
We complete the proof of through a simple substitution in Lemma \ref{lem: regret_uninformed}.
\end{proof}
\end{lemma}

To complete the proof of Theorem \ref{thm: bandit_lower} we can use Lemma 20 from \cite{Jaksch2010}.

\begin{proposition}[Bound on the KL divergence]
\label{prop: bound_kl}
For any $0 \le \delta \le \frac{1}{2}$ and $\epsilon \le 1 - 2 \delta$ we have
$$ \delta \log_2 \left(\frac{\delta}{\delta+\epsilon} \right) + (1 - \delta) \log_2\left( \frac{1 - \delta}{1 - \delta - \epsilon} \right) \le \frac{\epsilon^2}{\delta \log(2)}. $$

\end{proposition}

We combine Proposition \ref{prop: bound_kl} with Lemma \ref{lem: regret_kl} to say,
\begin{eqnarray*}
    \max_a \overline{r}^*(a) T - \Exp \left[ \sum_{t=1}^T  \overline{r}^*(\tilde{a}_t) \right]
    &\ge& \epsilon T \left(1 - \frac{1}{A} - \sqrt{\frac{\epsilon^2}{2 \delta} \frac{T}{A}} \right) \text{ for all } \epsilon\\
    &\ge& \sqrt{\frac{\delta A }{8T}} T \left(1 - \frac{1}{A} - \frac{1}{4} \right) \text{ by setting } \epsilon^2 = \frac{\delta A}{8 T} \\
    &\ge& \frac{1}{12}\sqrt{\delta AT}.
\end{eqnarray*}
We can choose $\delta = 0.25$ to complete the proof of Theorem \ref{thm: bandit_lower}.
We note that better constants are available through a more careful analysis, but this is not our focus in this work.
\qed

\section{Reinforcement learning}

In this section we will work to extend the lower bound arguments from bandits to reinforcement learning with $S \ge 2$.
As in common in the literature, we will begin with a simple two state MDP with known rewards and unknown transitions \cite{Jaksch2010,Bartlett2009,dann2015sample}.
It is relatively straightforward to extend this flavour of result to MDPs with $S > 2$ simply by concatenating $\lceil S/2 \rceil$ copies of these smaller systems.

State $0$ gives a reward of $0$ and state $1$ gives a reward of $1$.
All actions from the state 0 follow the same law $P(0,a) = (1-\delta_0, \delta_0)$.
In state 1 $P(1, a) = (\delta_1, 1-\delta_1)$ for all actions apart from $P(1, a^*) = (\delta_1 - \epsilon, 1 - \delta_1 + \epsilon)$.
For this simple MDP we will distinguish policies in terms of their action upon $s=1$, since this is the only action which can influence the evolution of the MDP.

\begin{figure}[h!]
    \centering
    \captionsetup{justification=centering,margin=1cm}
    \includegraphics[width=0.65\textwidth]{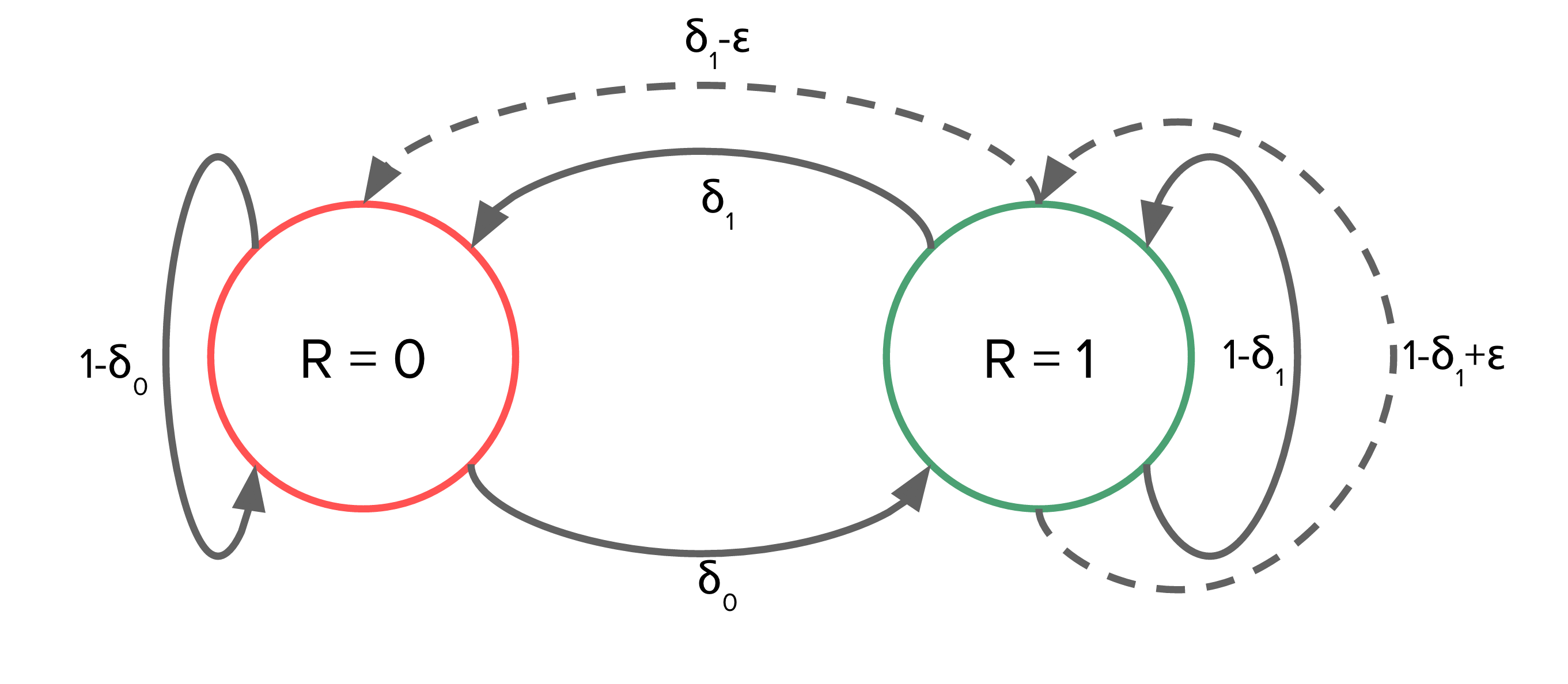}

    \caption{A two state MDP which is hard to learn. \\
             Dotted lines distinguish the unique optimal policy.}
    \label{fig: lower bound}
\end{figure}

We define $\theta_1 := \frac{\delta_0}{\delta_0 + \delta_1}$ to be the average expected reward under the policy $a \neq a^*$.
For convenience we write $\delta^*_1 := \delta_1 - \epsilon$ for the distinguished optimal action and correspondingly $\theta^*_1 := \frac{\delta_0}{\delta_0 + \delta^*_1}$ for the average expected reward under the optimal policy $a^*$.

\subsection{Sketch at REGAL-style lower bounds}
\label{sec: regal_bound}

In this section we present a quick overview of the style of argument that attempts to solidify the lower bound of Theorem 6 in \cite{Bartlett2009}.
We assume that $\delta_0 \ge \delta_1$ to bound the difference in optimal value,
\begin{eqnarray}
\label{eq: mdp_step_regret}
    \theta^*_1 - \theta_1 &=&
    \frac{\delta_0}{\delta_0 + \delta_1 - \epsilon} - \frac{\delta_0}{\delta_0 + \delta_1} \nonumber \\
    &=& \frac{\delta_0 \epsilon}{(\delta_0+\delta_1)(\delta_0 +\delta_1 - \epsilon)} \nonumber \\
    &>& \frac{\delta_0 \epsilon}{(\delta_0 +\delta_1)^2}
    > \frac{\delta_0 \epsilon}{(2 \delta_0 )^2} = \frac{\epsilon}{4 \delta_0}.
\end{eqnarray}
Broadly speaking, this indicates that the agent should obtain expected regret $\Omega(\epsilon / \delta_0)$ every timestep it selects action $a_t \neq a^*$ whilst in state $s=1$.
All other actions in any other state produce zero regret.
We now note that the problem described by Figure \ref{fig: lower bound} is quite similar to the bandit example from Section \ref{sec: mab}.
The difference here is that actions of the suboptimal arm $a \neq a^*$ give expected regret $O(\frac{\epsilon}{\delta_0})$, rather than $\epsilon$.

The arguments we present in this section can be thought of as an attempt to make the sketch proof for Theorem 6 of \cite{Bartlett2009} more explicit, if not entirely rigorous.
Our arguments will follow the same structure as Section \ref{sec: mab}:
we consider an auxilliary MDP where the optimal action $a^*$ has been replaced by another action with identical transition dynamics.
We will write $\tilde{a}_t$ for the actions which are taken by this uninformed policy and $\tilde{\Hc}_t$ for the uninformative history that it generates.
We begin with a result of a similar flavour to Lemma \ref{lem: regret_kl}.

\begin{lemma}[Regret of an uninformed agent]
\label{lem: mdp_regret_uninformed}
\hspace{0.000001mm} \newline
In the environment of Figure \ref{fig: lower bound}, for all $\delta, \epsilon > 0$ and all learning algorithms $\pi$,
$$ \max_a \overline{r}^*(a) T - \Exp \left[ \sum_{t=1}^T  \overline{r}^*(\tilde{a}_t) \right] \ge \theta_1 \frac{\epsilon}{4 \delta_0} T \left(1 - \frac{1}{A}\right)$$

\begin{proof}
We note that the uninformed agent can only incur regret when it makes a sub-optimal decision, which is only possible in state $s=1$.
The proportion of the time the agent spends in state $s=1$ is lower bounded by $\theta_1$.
The regret for any sub-optimal decision while in state $s=1$ is at least $\frac{\epsilon}{4 \delta_0}$ by \eqref{eq: mdp_step_regret}.
We follow the arguments from Lemma \ref{lem: regret_uninformed} to obtain our desired result.
\end{proof}
\end{lemma}

We now note that the problem of learning a 2-state transition function is equivalent to estimating a Bernoulli reward.
Therefore, we can use Lemma \ref{lem: mdp_regret_uninformed} in place of Lemma \ref{lem: regret_uninformed} and repeat a similar argument to the proof of Theorem \ref{thm: bandit_lower} for multi-armed bandits.
At a high level we can bound the regret of any agent in terms of the deviation in KL from the distribution of the uninformed agent.
For $\epsilon$ small, and over a short enough time window $T$, the distribution of actions chosen by the learning algorithm cannot differ significantly from the actions chosen from the uninformative system.
As such, using Pinsker's inequality, the resulting \textit{regret} from any learning algorithm cannot differ significantly from that of the uninformed algorithm.

To make this argument explicit, we use Lemma \ref{lem: kl_uninformed} and Lemma \ref{lem: regret_kl} together with Proposition \ref{sec: prob form} and optimize over the resulting bound over $\epsilon$.
That is to say, for any learning algorithm $\pi$,
\begin{eqnarray}
\label{eq: regal_lb}
    \theta^*_1 T - \Exp \left[ \sum_{t=1}^T  \overline{r}^*(a_t) \right]
    &\ge&  \theta_1 \frac{\epsilon}{4\delta_0} T \left( 1 - \frac{1}{A} -  \sqrt{\frac{1}{2} d_{KL}(\tilde{P}(z_1^T) , P(z_1^T))} \right) \nonumber \\
    &\ge& \theta_1 \frac{\epsilon}{4\delta_0} T
    \left( 1 - \frac{1}{A} - \sqrt{\frac{1}{2} \frac{\epsilon^2}{\delta_1} \frac{\theta_1 T}{ A}} \right) \text{ for all } \epsilon \nonumber \\
    &\ge& \frac{1}{4} \frac{\epsilon \theta_1 T}{\delta_0} \left(1 - \frac{1}{A} - \sqrt{\frac{\epsilon^2 \theta_1 T}{2 \delta_1 A}}\right) \nonumber \\
    &\ge& \frac{1}{4} \cdot \sqrt{\frac{\delta_1 A}{8 \theta_1 T}} \cdot \frac{\theta_1 T}{\delta_0} \left(1 - \frac{1}{A} - \frac{1}{4} \right) \text{ setting } \epsilon = \sqrt{\frac{\delta_1 A}{8 \theta_1 T}} \nonumber \\
    &\ge& \frac{1}{32 \sqrt{2}} \sqrt{\frac{\delta_1 \theta_1}{\delta_0^2} A T}.
\end{eqnarray}

Now, we are left with a problem to complete the argument for Theorem 6 from REGAL.
We introduce the notation, $T^M_\mu(s,s')$ for the expected number of timesteps to get from state $s$ to $s'$ in MDP $M$ under policy $\mu$.
The one-way diameter of an MDP is defined
\begin{equation}
    D_{\rm ow}(M) := \max_s \min_{\mu} T^M_\mu(s, \overline{s}), \text{ where } \overline{s} \text{ is any state with optimal value bias.}
\end{equation}
The claim in Theorem 6 of REGAL is that, for any learning algorithm $\pi$ there exists and MDP $M$ such that ${\rm Regret}(T, \pi, M^*) \ge c_0 D_{\rm ow} \sqrt{SAT}$ for some $c_0 > 0$.

From construction of the MDP in Figure \ref{fig: lower bound} it is clear that $D_{\rm ow} = \frac{1}{\delta_0}$, since the only state with optimal value bias is $s=1$ and the expected time from $s=0$ to $s=1$ is $\frac{1}{\delta_0}$.
We now examine behaviour of the remaining free parameters using the definition $\theta_1 = \delta_0 / (\delta_0 + \delta_1):$
\begin{eqnarray*}
    \sqrt{\frac{\delta_1 \theta_1}{\delta_0^2}}
    &=& D_{\rm ow} \sqrt{\delta_1 \theta_1} \\
    &=& D_{\rm ow} \sqrt{\frac{\delta_1 / D_{\rm ow}}{\delta_1 + 1/D_{\rm ow}}} \\
    &=& \sqrt{\frac{D_{\rm ow} }{1 + \frac{1}{\delta_1 D_{\rm ow}}}}
    = O(\sqrt{D_{\rm ow}}) \text{ for any choice of } \delta_1 > 0 .
\end{eqnarray*}

This completes the demonstration that the standard proof techniques for lower bounds do not address the problems in the proof REGAL Theorem 6.
In fact, we are only able to establish a lower bound $\Omega( \sqrt{ D_{\rm ow} SAT})$ and not $\Omega( D_{\rm ow} \sqrt{ SAT})$ as \cite{Bartlett2009} had claimed.
Further, these bounds are actually weaker than the established results in \cite{Jaksch2010} $\Omega( \sqrt{ D SAT})$, where $D(M) := \max_{s, s'} \min_{\mu} T^M_\mu(s, s') \ge D_{\rm ow}$ is the diameter of the MDP.

\subsection{Where do the lower bounds lie?}

The arguments in Section \ref{sec: regal_bound} show that existing machinery is not sufficient to establish a proof of Theorem 6 in \cite{Bartlett2009}.
In light of this we suggest that this published result be considered a conjecture, rather than an established theorem.
In this note we present another alternative conjecture, that the results of Theorem 6 in \cite{Bartlett2009} are not correct.
The spirit of this conjecture is similar to Conjecture 1 of \cite{osband2016posterior} given for finite horizon MDPs.

\begin{conjecture}[Tight lower bounds for regret]
\label{conj: tight_lower}
\hspace{0.00001mm} \newline
The lower bounds of \cite{Jaksch2010} $\Omega\left( \sqrt{DSAT} \right)$ are unimprovable in the sense that there exists some learning algorithm $\pi$ such that, for any MDP $M^*$ and any $\delta >0 $
\begin{equation}
    {\rm Regret}(T, \pi, M^*) = \tilde{O}\left( \sqrt{DSAT} \right),
\end{equation}
with probability at least $1 - \delta$.
\end{conjecture}

\subsubsection{What is wrong the REGAL lower bound?}
\label{sec: bad_regal}

In order for Conjecture \ref{conj: tight_lower} to be true, the sketched proof in \cite{Bartlett2009} must be false.
Although the arguments of Section \ref{sec: regal_bound} show that this proof is not yet rigorous, they do not pinpoint any step of the appealing sketched argument which is incorrect.
However, we will now present an intuitive argument for what may be going wrong in the sketched proof:
\vspace{-3mm}
\begin{itemize}
    \item For every timestep $t$ in state $s=1$ the worst possible decision the agent could make will contribute regret $O(D{\rm ow})$ in terms of the \textit{value}.
    The proposed sketch proof argues that the agent effectively incurs this regret every timestep until it learns the optimal arm.

    \item If we measure regret in terms of actual shortfall in the instantaneous regret $\lambda^*_* - r_t$ must be bounded $O(1)$ per timestep.
    The bad decisions in state $s=1$ are just worth $O(D_{\rm ow})$ \textit{value} because it might lead to $O(D_{\rm ow})$ of these $O(1)$ instantaneous regret steps to occur in a row.

    \item Alternatively, we might think of regret in terms of the future \textit{value} $O(D_{\rm ow}$ which a bad decision at $s=1$ may be worth - this is the argument that REGAL uses \cite{Bartlett2009}.
    However, if we do this then that means this bad decision must be followed by $O(D_{\rm ow})$ timesteps in which we count no additional regret.
\end{itemize}

At the moment, the argument for Theorem 6 in \cite{Bartlett2009} is doing a type of double-counting for regret.
It assigns the maximum $O(D_{\rm ow}(M^*))$ regret in terms of \textit{value} at each \textit{timestep}.
However, this analysis ignores that for every one of these bad actions there will be $O(D_{\rm ow})(M^*)$ periods of time within $s=0$ where, in terms of the value shortfall, these actions will not incur further regret than has been counted already.


\subsubsection{Comparison to existing tight PAC bounds}
\label{sec: compare_pac}

Another piece of tangentially supporting evidence for Conjecture \ref{conj: tight_lower} comes from the recent PAC-analysis for finite horizon MDPs \cite{dann2015sample}.
The problem formulation given by this paper differs from \cite{Bartlett2009} in several ways, but they produce an algorithm LUCFH which matches upper and lower bounds for the horizon $H$ in finite horizon MDPs.
In finite horizon MDPs, the horizon $H$ is an upper bound on $D_{\rm ow}$.
A similar flavour of result is available in discounted MDPs \cite{lattimore2012pac} where the horizon $H$ is replace with an equivalent timeframe $H = \tilde{O} \left(\frac{1}{1-\gamma}\right)$.

The analysis for LUCFH in finite horizon MDPs implies that the number of \textit{episodes} required for $\epsilon$-optimal \textit{episodes} is $\Theta ( \frac{H^2}{\epsilon^2} )$, where we view all variables other than $H$ and $\epsilon$ as fixed.
According to their definition, this would imply $\Theta ( \frac{H^3}{\epsilon^2} )$ \textit{timesteps} until $\epsilon$-optimal episodes, which is roughly equivalent to $\Theta ( \frac{H}{\epsilon^2} )$ \textit{timesteps} until $\epsilon$-optimal timesteps.

At a high level the algorithm and analysis from \cite{dann2015sample} leverages the sort of phenomenon we describe in Section \ref{sec: bad_regal}.
This essential argument is refined and made more rigorous through the Bellman equation for local variance, first used in \cite{lattimore2012pac}.
It is not generally possible to go from PAC bounds to regret guarantees, however, the spirit of previous analyses and comparable results suggest that the tight bounds $\Theta ( \frac{H}{\epsilon^2} )$ \textit{timesteps} until $\epsilon$-optimal timesteps are suggestive of a tight regret scaling $\Theta(\sqrt{HT})$.

\section{Conclusion}

This technical note aims to clarify the current state of lower bounds for regret in reinforcement learning.
We reproduce a clear step by step argument for the lower bound on regret given in \cite{Bartlett2009}.
We show that, using standard machinery, this leads to a provable lower bound $\Omega(\sqrt{D_{\rm ow}SAT})$ and currently there is no proof available for the bound $\Omega( D_{\rm ow} \sqrt{SAT})$ as conjectured in that earlier work.
To stimulate thinking on this topic, we present Conjecture \ref{conj: tight_lower}, that the lower bound $\Omega(\sqrt{D_{\rm ow} SAT})$ is in fact unimprovable.
Definitively proving these results one way or another is an exciting area for future research.

\section*{Acknowledgements}

We would like to thank the authors of \cite{Bartlett2009} for their help and dialogue in the discussion of these delicate technical issues.
We would also like to thank Daniel Russo for the many hours of discussion and analysis spent in the office on issues like these.

{
\small
\bibliographystyle{plainnat}
\bibliography{reference}
}

\end{document}